\documentclass{sig-alternate}

\usepackage{url}

\begin{document}



\doi{}

\isbn{}



%

\title{Direct-Manipulation Visualization of Deep Networks}
%
%
%
%
%

\numberofauthors{5} 
%
\author{
%
%
\alignauthor
Daniel Smilkov\\
       \affaddr{Google, Inc.}\\
       \affaddr{5 Cambridge Center, Cambridge MA, 02142}\\
       \email{smilkov@google.com}
\alignauthor
Shan Carter\\
       \affaddr{Google, Inc.}\\
       \affaddr{1600 Amphitheater Parkway, Mountain View CA, 94043}\\
       \email{shancarter@google.com}
\alignauthor
D. Sculley\\
       \affaddr{Google, Inc.}\\
       \affaddr{5 Cambridge Center, Cambridge MA, 02142}\\
       \email{dsculley@google.com}
\and  
\alignauthor
Fernanda B. Vi\'{e}gas\\
       \affaddr{Google, Inc.}\\
       \affaddr{5 Cambridge Center, Cambridge MA, 02142}\\
       \email{viegas@google.com}
\alignauthor
Martin Wattenberg\\
       \affaddr{Google, Inc.}\\
       \affaddr{5 Cambridge Center, Cambridge MA, 02142}\\
       \email{wattenberg@google.com}
}
\date{17 May 2016}

\maketitle
\begin{abstract}
The recent successes of deep learning have led to a wave of interest from non-experts. Gaining an understanding of this technology, however, is difficult. While the theory is important, it is also helpful for novices to develop an intuitive feel for the effect of different hyperparameters and structural variations.
We describe TensorFlow Playground\footnote{\url{http://playground.tensorflow.org}}, an interactive, open sourced\footnote{\url{https://github.com/tensorflow/playground}} visualization that allows users to experiment via direct manipulation rather than coding, enabling them to quickly build an intuition about neural nets.
\end{abstract}

%
%


%
%

%
%



\section{Introduction}

Deep learning systems are currently attracting a huge amount of interest, as they see continued success in practical applications. Students who want to understand this new technology encounter two primary challenges.

First, the theoretical foundations of the field are not always easy for a typical software engineer or computer science student, since they require a solid mathematical intuition. It's not trivial to translate the equations defining a deep network into a mental model of the underlying geometric transformations.

Even more challenging are aspects of deep learning where theory does not provide crisp, clean explanations. Critical choices experts make in building a real-world system--the number of units and layers, the activation function, regularization techniques, etc.--are currently guided by intuition and experience as much as theory. Acquiring this intuition is a lengthy process, since it typically requires coding and training many different working systems.

One possible shortcut is to use interactive visualization to help novices with mathematical and practical intuition. Recently, several impressive systems have appeared that do exactly this. Olah's elegant interactive online essays \cite{colahsblog} let a viewer watch the training of a simple classifier, providing a multiple perspectives on how a network learns a transformation of space. Karpathy created a Javascript library \cite{convnetjs} and provided a series of dynamic views of networks training, again in a browser. Others have found beautiful ways to visualize the features learned by image classification nets \cite{zhou}, \cite{zeiler}.

Taking inspiration from the success of these examples, we created the TensorFlow Playground. As with the work of Olah and Karpathy, the Playground is an in-browser visualization of a running neural network. However, it is specifically designed for experimentation by direct manipulation, and also visualizes the derived ``features'' found by every unit in the network simultaneously. The system provides a variety of affordances for rapidly and incrementally changing hyperparameters and immediately seeing the effects of those changes, as well as for sharing experiments with others.

\begin{figure*}[ht]
\vskip 0.2in
\begin{center}
\centerline{\includegraphics[width=6.5in]{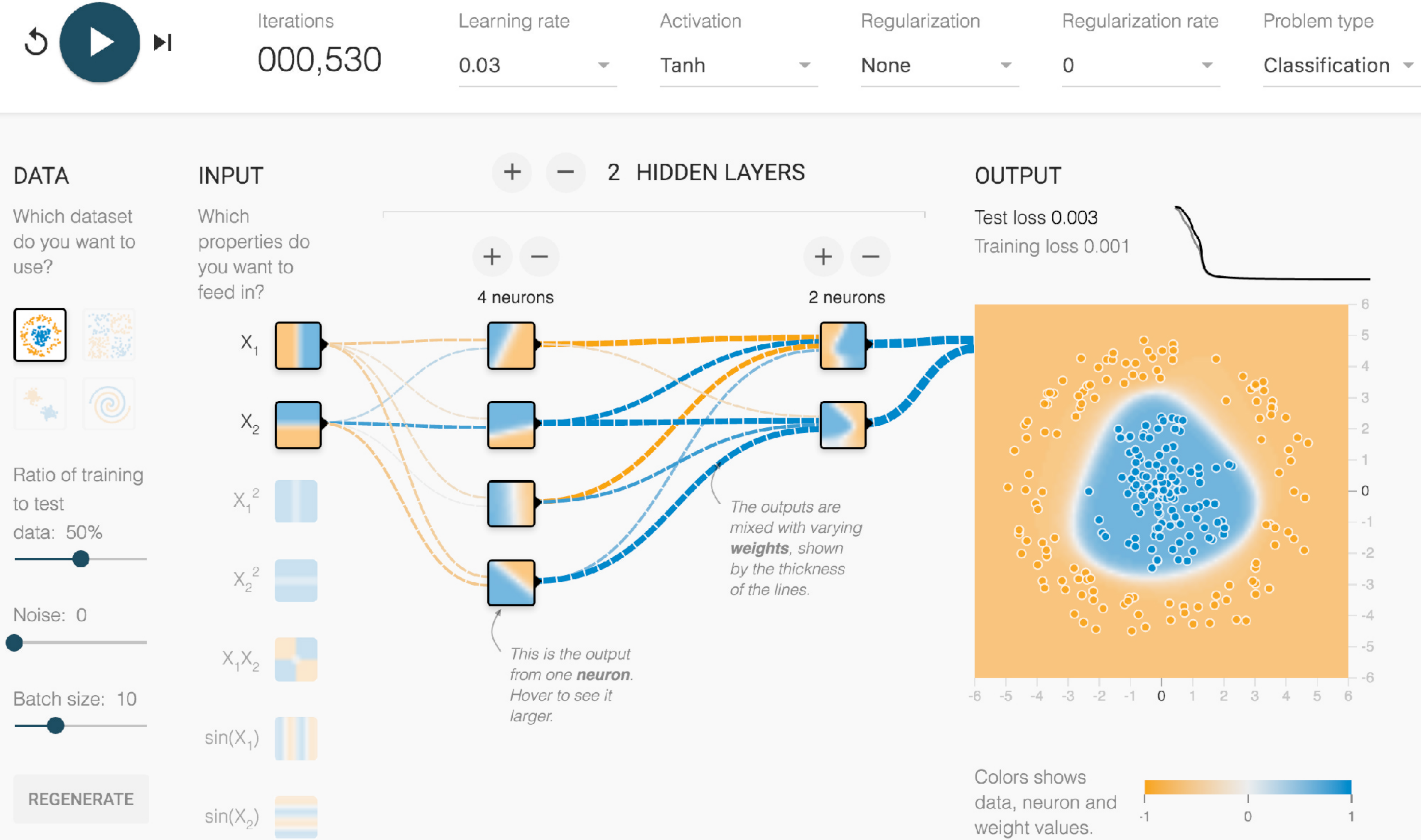}}
\caption{TensorFlow Playground. This network is, roughly speaking, classifying data based on distance to the origin. Curves show weight parameters, with thickness denoting absolute magnitude and color indicating sign. The feature heatmaps for each unit show how the classification function (large heatmap at right) is built from input features, then near-linear combinations of these features, and finally more complex features. At upper right is a graph showing loss over time. At left are possible features; $x_1$ and $x_2$ are highlighted, while other mathematical combinations are faded to indicate they should not be used by the network.}
\label{playground-initial}
\end{center}
\vskip -0.2in
\end{figure*}

\section{TensorFlow Playground: Visualization}

The structure of the Playground visualization is a standard network diagram. The visualization shows a network that is designed to solve either classification or regression problems based on two abstract real-valued features, $x_1$ and $x_2$, which vary between -1 and 1. Input units, representing these features and various mathematical combinations, are at the left. Units in hidden layers are shown as small boxes, with connections between units drawn as curves whose color and width indicate weight values. Finally, on the right, a visualization of the output of the network is shown: a square with a heatmap showing the output value of the single unit that makes up the final layer of the network. When the user presses the "play" button, the network begins to train.

There is a new twist in this visualization, however. Inside the box that represents each unit is a heatmap that maps the unit's response to all values of $(x_1, x_2)$ in a square centered at the origin. As seen in Figure~\ref{playground-initial}, this provides a quick geometric view of how the network builds complex features from simpler ones. For example, in the figure the input features are simply $x_1$ and $x_2$, which themselves are represented by the same type of heatmap. In the next layer, we see units that correspond to various linear combinations, leading to a final layer with more complicated non-linear classifiers. Moving the mouse over any of these units projects a larger version of the heatmap, on the final unit, where it can be overlaid with input and test data.

\begin{figure*}[ht]
\vskip 0.2in
\begin{center}
\centerline{\includegraphics[width=6.5in]{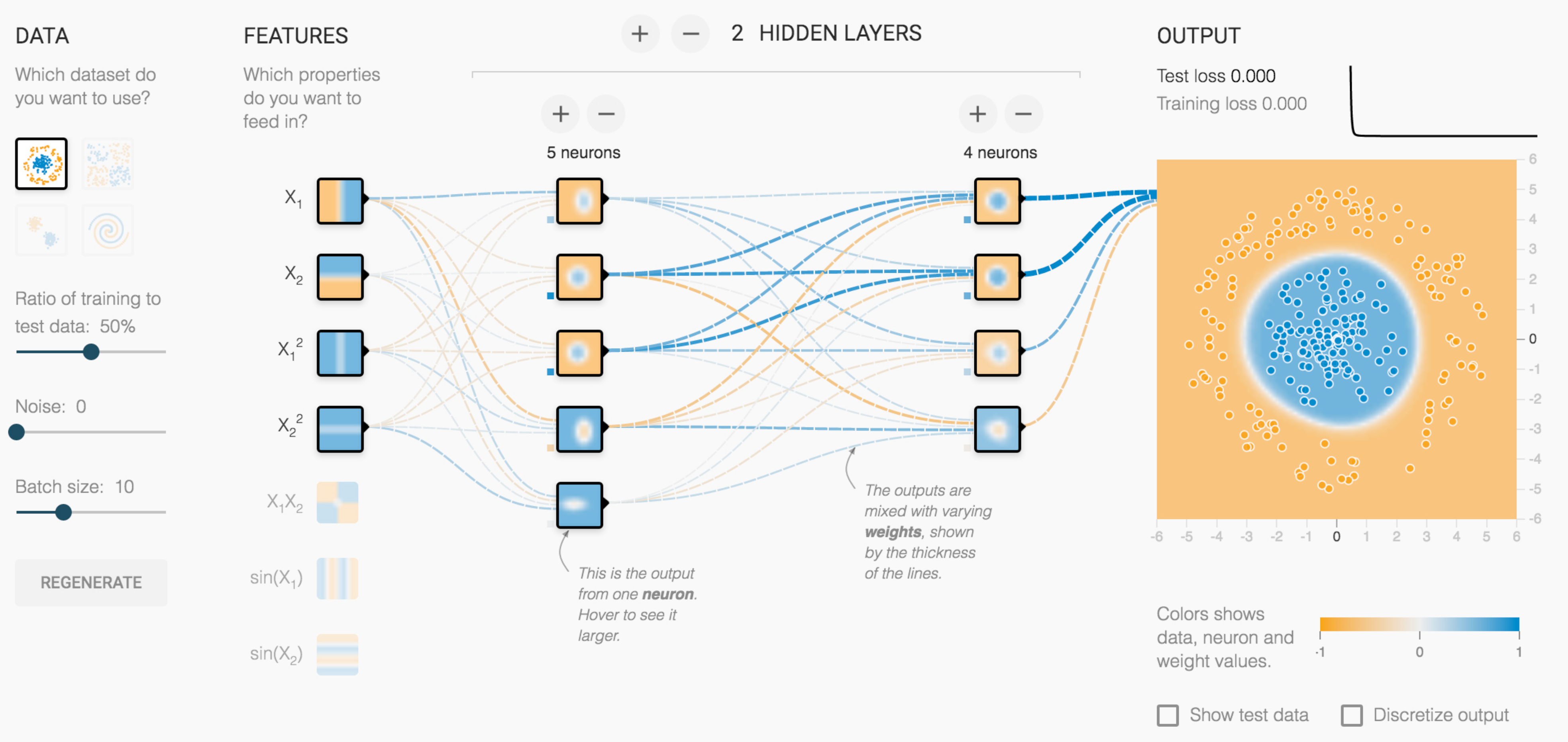}}
\caption{A network architecture with redundant layers and units. Several units in the first hidden layer have already essentially learned to classify the data, as seen by inspecting the in-network activation visualizations.}
\label{playground-redundant}
\end{center}
\vskip -0.2in
\end{figure*}

\begin{figure*}[ht]
\vskip 0.2in
\begin{center}
\centerline{\includegraphics[width=6.5in]{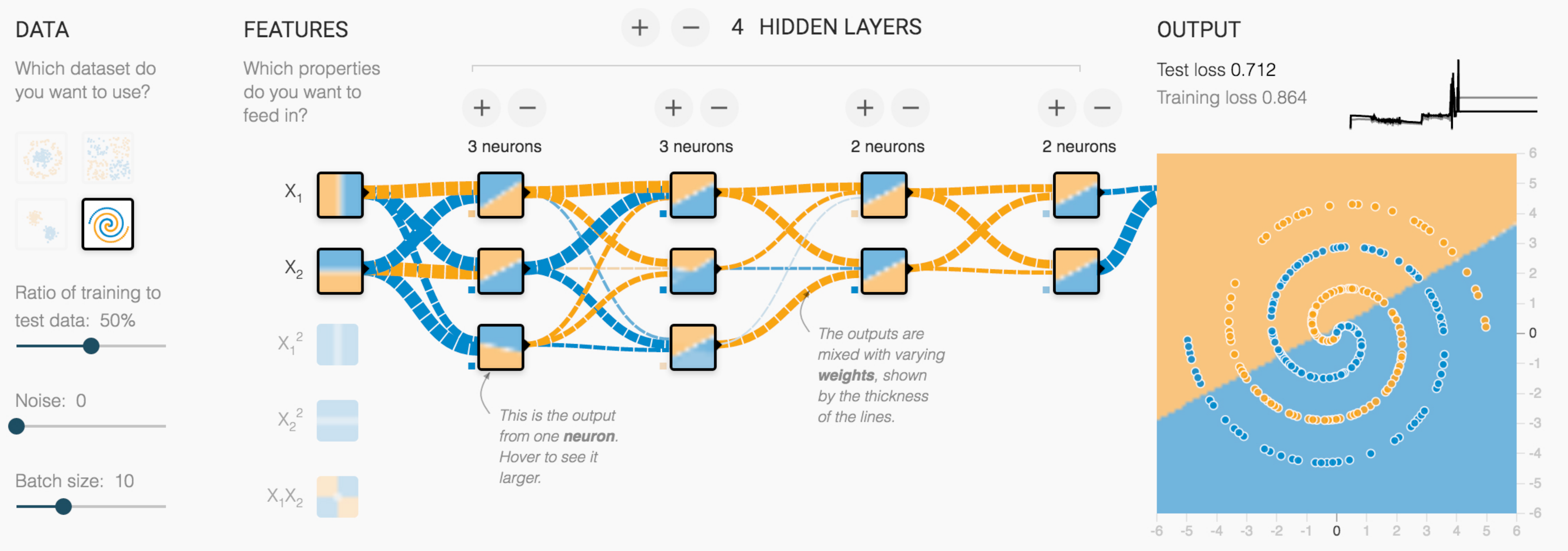}}
\caption{This network has completely failed to classify the data, even after many epochs. The high-contrast activation visualizations and thick weight connections hint at a systemic problem. This diagram was the result of setting the learning rate to the maximum speed.}
\label{playground-failed}
\end{center}
\vskip -0.2in
\end{figure*}

The activation heatmaps help users build a mental model of the mathematics underlying deep networks. For many configurations of the network, after training there is an obvious visual progression in complexity across the network. In these configurations, viewers can see how the first layer of units (modulo activation function, acting as linear classifiers) combine to recognize clearly nonlinear regions. The heatmaps also help viewers understand the different effects of various activation functions. For example, there is a clear visual difference in the effect of ReLU and $\tanh$ functions. Just as instructive, however, are suboptimal combinations of architecture and hyperparameters. Often when there are redundant units (Figure~\ref{playground-redundant}), it is easy to see that units in intermediate layers have actually learned the classifier perfectly well and that many other units have little effect on the final outcome. In cases where learning is simply unsuccessful, the viewer will often see weights going to zero, and that there is no natural progression of complexity in the activation heatmaps (Figure~\ref{playground-failed}).

The visualization is implemented in JavaScript using d3.js\cite{2011-d3}. It is worth noting that for the neural network computation, we are not using the TensorFlow library\cite{tensorflow2015-whitepaper} since we needed the whole visualization to run in the browser. Instead, we wrote a small library\footnote{\url{https://github.com/tensorflow/playground/blob/master/nn.ts}} that meets the demands of this educational visualization.

\begin{figure*}[ht]
\vskip 0.2in
\begin{center}
\centerline{\includegraphics[width=6.5in]{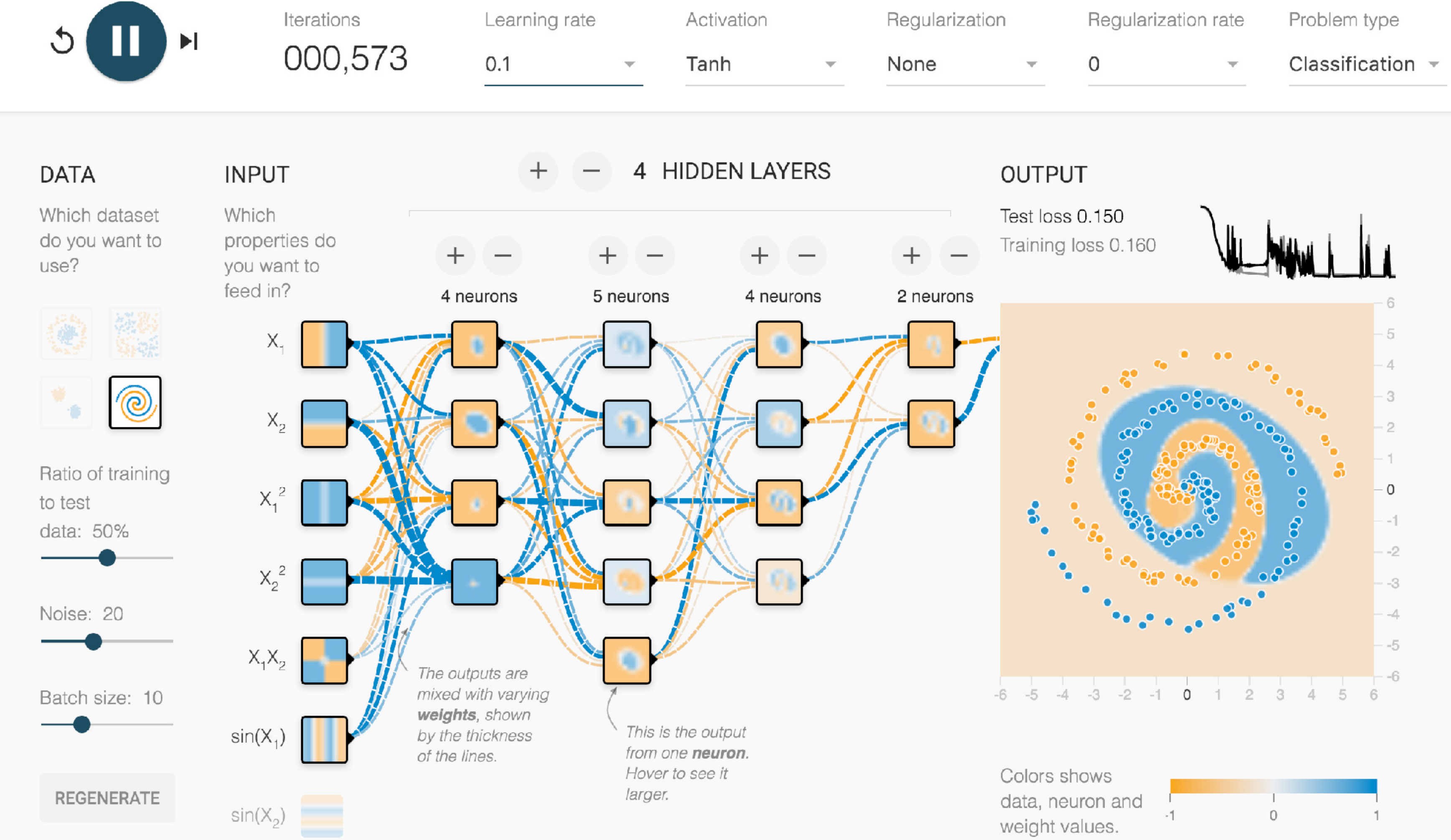}}
\caption{A complex configuration of TensorFlow Playground, in which a user is attempting to find hyperparameters that will allow the classification of spiral data. Many possible feature combinations have been activated.}
\label{playground-complex}
\end{center}
\vskip -0.2in
\end{figure*} 

\section{Affordances for Education and Experimentation}

The real strength of this visualization is its interactivity, which is especially helpful for gaining an intuition for the practical aspects of training a deep network. The Playground lets users make the following choices of network structure and hyperparameters:

\begin{itemize}
\item Problem type: regression or classification
\item Training data: a choice of four synthetic data sets, from well-separated clusters to interleaved "swiss roll" spirals.
\item Number of layers
\item Number of units in each layer
\item Activation function
\item Learning rate
\item Batch size
\item Regularization: $L^1$, $L^2$, or none
\item Input features: in addition to the two real-valued features $x_1$ and $x_2$, the Playground allows users to add some simple algebraic combinations, such as $x_1 x_2$ and $x_1^2$.
\item Noise level for input data
\end{itemize}

These particular variations were chosen based on experience teaching software engineers how to use neural networks in their applications, and are meant to highlight key decisions that are made in real life. They are also meant to be easily combined to support particular lessons. For instance, allowing users to add algebraic combinations of the two primary features makes it easy to show how a linear classifier can do "non-linear" tasks when given non-linear feature combinations.

The user interface is designed to make these choices as easy to modify as possible. The standard definition of direct manipulation is that changes should be "rapid, incremental and reversible" \cite{shneiderman19931}. Allowing fast, smooth changes to variables helps build intuition for their effects. Reversibility encourages experimentation: indeed, we chose as our tagline for the visualization, "You can't break it. We promise."

Additional aspects of the visualization make it well-suited to education. We have found that the smooth animation engages users. It also lends itself to a good ``spectator experience'' \cite{reeves2005designing}, drawing students in during presentations. We have seen onlookers laugh and even gasp as they watch a network try and fail to classify the spiral data set, for example. Although animation has not always been found to be helpful in educational contexts, simulations are one case where there is good evidence that it is beneficial \cite{betrancourt2005animation}.

One particularly important feature is the ability to seamlessly bookmark \cite{viegas2007manyeyes} a particular configuration of hyperparameters and structure. As the user plays with the tool, the URL in the browser dynamically updates to reflect its current state. If the user (or a teacher preparing a lesson plan) finds a configuration they would like to share with others, they need only copy the URL. Additionally, using the checkboxes below the visualization, each UI component can be hidden, making it easy to repurpose the interface.

We have found this bookmarking capability invaluable in the teaching process. For example, it has allowed us to put together tutorials in which students can move, step by step, through a series of lessons that focus on particular aspects of neural networks. Using the visualization in these ``living lessons'' makes it straightforward to create a dynamic, interactive educational experience.

\section{Conclusion and Future Work}

The TensorFlow Playground illustrates a direct-manipulation approach to understanding neural nets. Given the importance of intuition and experimentation to the field of deep learning, the visualization is designed to make it easy to get a hands-on feel for how these systems work without any coding. Not only does this extend the reach of the tool to people who aren't programmers, it provides a much faster route, even for coders, to try many variations quickly. By playing with the visualization, users have a chance to build a mental model of the mathematics behind deep learning, as well as develop a natural feeling for how these networks respond to tweaks in architecture and hyperparameters.

In addition to internal success with the tool, we have seen a strong positive reaction since it has been open-sourced. Besides general positive comments, we have seen interesting, playful interactions. On one Reddit thread, for example, people competed to find a way to classify the spiral data, posting screenshots of their successful configurations. This suggests that the tool is instigating a vibrant social reaction to the visualization.

Since the launch of TensorFlow Playground, we have seen many suggestions for extensions. Affordances for many other structural variations and hyperparameters could be added; for instance, a common request is for an option to see the effect of dropout. Architectures such as convolutional nets and LSTMs could also be illuminated through direct manipulation techniques. Our hope is that, as an open-source project, the Playground will be extended to accommodate many such ideas. More broadly, the ideas of visualization, direct manipulation, and shareability that we have used may prove useful in explaining other aspects of deep learning besides network structure and hyperparameters.

A further question is whether this same direct-manipulation environment can be extended to help researchers as well as students. While there are obvious technical obstacles--breaking new ground often requires large data sets and computational resources beyond what a browser offers--it may be possible to create minimal "research playgrounds" that yield insights and allow rapid experimentation.

%
\bibliographystyle{abbrv}
\bibliography{main}

\begin{thebibliography}{10}

\bibitem{tensorflow2015-whitepaper}
M.~Abadi, A.~Agarwal, P.~Barham, E.~Brevdo, Z.~Chen, C.~Citro, G.~S. Corrado,
  A.~Davis, J.~Dean, M.~Devin, S.~Ghemawat, I.~Goodfellow, A.~Harp, G.~Irving,
  M.~Isard, Y.~Jia, R.~Jozefowicz, L.~Kaiser, M.~Kudlur, J.~Levenberg,
  D.~Man\'{e}, R.~Monga, S.~Moore, D.~Murray, C.~Olah, M.~Schuster, J.~Shlens,
  B.~Steiner, I.~Sutskever, K.~Talwar, P.~Tucker, V.~Vanhoucke, V.~Vasudevan,
  F.~Vi\'{e}gas, O.~Vinyals, P.~Warden, M.~Wattenberg, M.~Wicke, Y.~Yu, and
  X.~Zheng.
\newblock {TensorFlow}: Large-scale machine learning on heterogeneous systems,
  2015.
\newblock Software available from tensorflow.org.

\bibitem{betrancourt2005animation}
M.~Betrancourt.
\newblock The animation and interactivity principles in multimedia learning.
\newblock {\em The Cambridge handbook of multimedia learning}, pages 287--296,
  2005.

\bibitem{2011-d3}
M.~Bostock, V.~Ogievetsky, and J.~Heer.
\newblock D3: Data-driven documents.
\newblock {\em IEEE Trans. Visualization \& Comp. Graphics (Proc. InfoVis)},
  2011.

\bibitem{convnetjs}
A.~Karpathy.
\newblock Convnetjs: Deep learning in your browser.

\bibitem{colahsblog}
C.~Olah.
\newblock colah's blog.

\bibitem{reeves2005designing}
S.~Reeves, S.~Benford, C.~O'Malley, and M.~Fraser.
\newblock Designing the spectator experience.
\newblock In {\em Proceedings of the SIGCHI conference on Human factors in
  computing systems}, pages 741--750. ACM, 2005.

\bibitem{shneiderman19931}
B.~Shneiderman.
\newblock 1.1 direct manipulation: a step beyond programming languages.
\newblock {\em Sparks of innovation in human-computer interaction}, 17:1993,
  1993.

\bibitem{viegas2007manyeyes}
F.~B. Viegas, M.~Wattenberg, F.~Van~Ham, J.~Kriss, and M.~McKeon.
\newblock Manyeyes: a site for visualization at internet scale.
\newblock {\em Visualization and Computer Graphics, IEEE Transactions on},
  13(6):1121--1128, 2007.

\bibitem{zeiler}
M.~D. Zeiler and R.~Fergus.
\newblock Visualizing and understanding convolutional networks.
\newblock In {\em Computer vision--ECCV 2014}, pages 818--833. Springer, 2014.

\bibitem{zhou}
B.~Zhou, A.~Khosla, {\`{A}}.~Lapedriza, A.~Oliva, and A.~Torralba.
\newblock Object detectors emerge in deep scene cnns.
\newblock {\em CoRR}, abs/1412.6856, 2014.

\end{thebibliography}

%
%
\end{document}